\documentclass[11pt]{article}
\usepackage{booktabs}
\usepackage{multirow}
\usepackage{amsmath}
\usepackage{subcaption}
\usepackage{float}
\usepackage[table]{xcolor}
\usepackage{hyperref}
\usepackage[preprint]{acl}

\usepackage{times}
\usepackage{latexsym}

\usepackage[T1]{fontenc}

\usepackage[utf8]{inputenc}

\usepackage{microtype}

\usepackage{inconsolata}

\usepackage{graphicx}
\usepackage{enumitem}
\usepackage{algorithm}
\usepackage{algorithmic}
\title{LIBERO-Occ: Evaluating and Improving Vision-Language-Action Models under Scene-Induced Occlusion via Viewpoint Imagination}

\author{
    Taishan Li$^{1}$\thanks{Equal contribution.}, Jiwen Zhang$^{1}$\footnotemark[1], Siyuan Wang$^{3}$\thanks{Corresponding author.}, Xuanjing Huang$^{1}$, Zhongyu Wei$^{1,2}$\footnotemark[2]\\
    $^{1}$Fudan University, Shanghai, China\\
    $^{2}$Shanghai Innovation Institute, Shanghai, China\\
    $^{3}$Chinese University of Hong Kong, Hong Kong, China\\
    \texttt{\{tsli23,jiwenzhang21\}@m.fudan.edu.cn;siyuanwang@cuhk.edu.hk;\{xjhuang,zywei\}@fudan.edu.cn}\\
}

\begin{document}
\maketitle
\begin{abstract}
Vision-Language-Action (VLA) models have achieved strong performance on standard manipulation benchmarks, but most evaluations assume that task-relevant objects are fully visible. This assumption often fails in realistic settings, where occlusion makes manipulation partially observable. In this paper, we study \textit{scene-induced occlusion} as a fundamental challenge for VLA models and introduce \textbf{LIBERO-Occ}, an occlusion-oriented extension of LIBERO. Experiments show that state-of-the-art VLAs suffer substantial performance degradation under occlusion. To address this issue, we propose \textbf{Viewpoint Imagination (VIM)}, which generates a complementary view from an occluded primary observation and conditions action prediction on both observed and imagined evidence. VIM improves robustness across task suites, occlusion types, and severity levels without requiring additional cameras at deployment time, suggesting that viewpoint imagination is a promising mechanism for perception completion in partially observable manipulation. Our benchmark and corresponding code are available at: \href{https://github.com/litsh/Libero-Occ}{https://github.com/litsh/Libero-Occ}.
\end{abstract}

\section{Introduction}

Recently, Vision-Language-Action (VLA) models have emerged as a promising paradigm for generalizable robotic manipulation. Some VLAs~\cite{Kim2025FineTuningVM, pmlr-v305-black25a, UniVLA} have achieved strong performance on standard manipulation benchmarks. However, most existing evaluation protocols implicitly assume that task-relevant objects are fully visible in the input observation. This assumption rarely holds in real-world manipulation environments, where occlusion is a common phenomenon~\cite{Guruprasad2024BenchmarkingVL, Zhang2026TacVLACT}.
Under occlusion, manipulation becomes a partially observable decision-making problem~\cite{Zhao2021HierarchicalPP, Xiao2019OnlinePF}, demanding the ability to infer unobserved scene states~\cite{Back2021UnseenOA, Miao2022SafeOM}.
As a result, a critical challenge for deployable VLA systems is whether they can operate when critical task-relevant evidence is not fully observable. 

\begin{figure}[t]
\centering
\setlength{\abovecaptionskip}{-2pt}
\includegraphics[width=\linewidth]{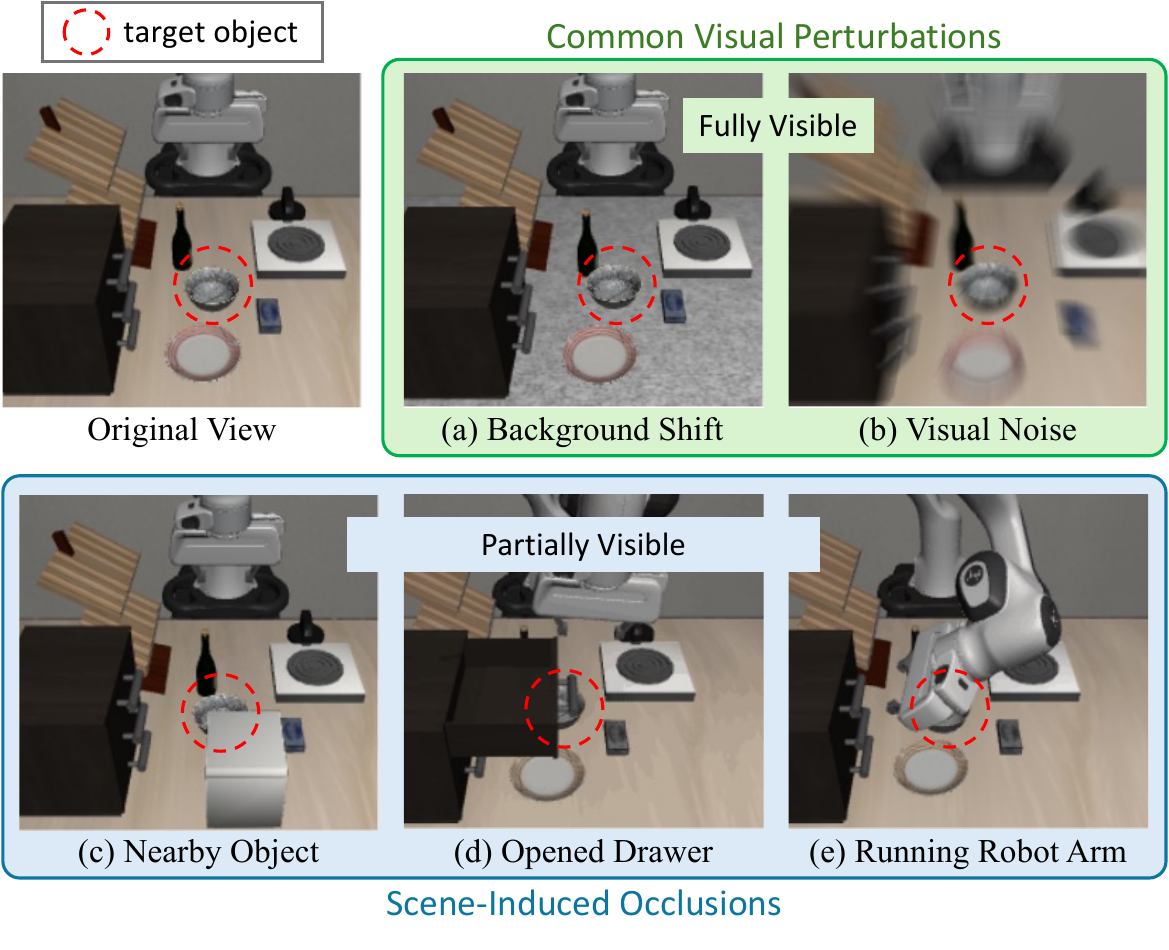}
\caption{
    \textbf{Comparison with common perturbations.}
    Task instruction: \emph{"open the top drawer and put the bowl inside"}.
    Unlike artificially-crafted perturbations, scene-induced occlusions remove task-relevant evidence from the observation and increase the task difficulty.
    }
    \label{fig:motivation}
\end{figure}

We refer to this problem as \textit{scene-induced occlusion}, an intrinsic challenge arising from the physical configuration and dynamics of the manipulation scene itself. As shown in Figure~\ref{fig:motivation}, the target object may be initially hidden behind nearby objects, or be covered by manipulated objects or robot arms during operation. Unlike common visual perturbations~\cite{robustVLA,libero-plus} that artificially alter observation appearance, scene-induced occlusion preserves scene texture while reducing the visibility of task-relevant evidence. Despite its practical importance, scene-induced occlusion remains underexplored in the VLA domain, making it unclear how robust existing models are under such partial observability.

A straightforward way to mitigate occlusion is to increase the physical visibility of the target object, either by introducing additional cameras~\cite{Bian2025VLALPAFLP} or by actively adjusting the camera viewpoint~\cite{Liu2026SaPaVeTA, Liu2026ActiveVLAIA}. Though effective, these approaches can be costly in real deployment, as additional cameras require calibration, maintenance, and appropriate placement; and active camera systems introduce extra hardware constraints. This raises a central question: can a VLA model recover missing task-relevant information without adjusting hardware?

Motivated by recent progress in embodied world models and generative modeling~\cite{cen2025worldvlaautoregressiveactionworld, UniVLA}, we explore \textbf{Viewpoint Imagination (VIM)} as a mechanism for perception completion. The key intuition is that generation can serve as a form of understanding: if a model can imagine a plausible complementary view from an occluded observation, it may recover spatial evidence that is hidden in the primary camera. We therefore propose a viewpoint imagination-based framework for VLA-based manipulation under scene-induced occlusion. Our framework is built upon a world model and aims to transform its generative prior into a perception-completion capability for manipulation. Given an occluded primary observation, the model first generates a complementary viewpoint that may reveal task-relevant information. It then predicts actions conditioned on both the observed image and the imagined view. The framework is trained in two stages: it first learns to generate a complementary viewpoint from the occluded observation, and then jointly optimizes viewpoint imagination and action prediction so that the generated visual evidence is aligned with downstream policy learning. At inference time, the method requires only the original camera observation, avoiding the need for additional hardware.

To systematically evaluate the influence of scene-induced occlusion in VLA manipulation, we construct LIBERO-Occ, an occlusion-oriented extension of the LIBERO benchmark~\cite{Liu2023LIBEROBK}. LIBERO-Occ introduces controlled scene-induced occlusions while preserving the original task semantics. 
Moreover, we include multiple occlusion severity levels, enabling reliable evaluation under different degrees of missing task-relevant information. Experiments show that existing VLA models suffer substantial performance degradation under scene-induced occlusion, while our proposed viewpoint imagination framework remains effective, demonstrating the feasibility of using imagined complementary views to improve manipulation robustness under partial observability.

In summary, our contributions are threefold:
\begin{itemize}[leftmargin=*,partopsep=0pt,topsep=0pt]
\setlength{\itemsep}{0pt}
\setlength{\parsep}{0pt}
\setlength{\parskip}{0pt}
    \item We identify scene-induced occlusion as an important and underexplored challenge for VLA.
    \item We introduce \textbf{LIBERO-Occ}, a benchmark for systematic evaluation under controlled occlusion types and severity levels.
    \item We propose \textbf{VIM}, a viewpoint imagination-based VLA framework that tackles scene-induced occlusion by inferring complementary visual evidence without modifying the hardware setup.
    
\end{itemize}

\section{LIBERO-Occ Benchmark}
To systematically evaluate scene-induced occlusion, we introduce LIBERO-Occ, an occlusion-oriented benchmark built upon LIBERO~\cite{Liu2023LIBEROBK}. LIBERO is a widely used robotic manipulation benchmark that covers four task suites---Spatial, Object, Goal, and 10---for evaluating generalization across layouts, objects, goals, and long-horizon tasks. LIBERO-Occ extends these suites by introducing physically grounded occlusions while preserving the original task semantics and executability.


\subsection{Occlusion Task Generation}
Constructing realistic scene-induced occlusions is non-trivial. Valid occluder placements must effectively block the target from the camera while jointly respecting scene geometry, target object location, and workspace layout, while the resulting task must remain physically executable.
To construct LIBERO-Occ at scale, we develop an automatic task generation pipeline that builds occluded task scenes from the original LIBERO tasks.

\paragraph{Step 1: Occlusion Target Identification.}
Given an original LIBERO task, we first parse its BDDL\footnote{
Behavior Domain Definition Language (BDDL) is the symbolic task specification format used by LIBERO.  It defines a manipulation task through involved objects, initial scene conditions, and goal predicates.
} specification to identify task-relevant entities as occlusion targets. These entities include manipulated objects, receptacles or goal regions. This step determines which objects or regions should be considered valid occlusion targets.

\paragraph{Step 2: View-aware Occluder Placement.}
To generate physical occlusions rather than image-space masks, we sample occluder placements in 3D scene space. Specifically, for each occlusion target, we calculate candidate placement locations along the camera-to-target ray from the primary view, so that the inserted occluder is likely to block the target in the observation. We then sample occluder objects from the LIBERO object library and filter out objects that semantically overlap with existing task objects, avoiding ambiguities that could change the intended task semantics.

\paragraph{Step 3: Occlusion Validity Verification.}
Each feasible candidate occluder is instantiated as an additional object in the scene and rendered in the simulator. To make sure the added occluder is physically plausible, we conduct three verification steps. We retain candidates only if they satisfy three constraints:
(1) \textit{Visibility check}: We verify the occlusion effect by rendering both the occluded scene and a reference scene with the occluder removed. We reject candidates that either fail to induce sufficient occlusion or make the target almost completely invisible and thus infeasible.
(2) \textit{Physical validity check}: We reject tasks in which the inserted occluder collides with existing objects after reset.
(3) \textit{Task executability check}: Finally, we replay the original demonstration in the occluded scene and retain only variants where the task can still be successfully completed. This step ensures that each retained task remains executable under the introduced occlusion, so that model failures can be attributed to occlusion-induced partial observability rather than invalid task construction.


\subsection{Benchmark Characterization}
LIBERO-Occ characterizes each occluded task along two axes: \textit{what} task-relevant information is hidden and \textit{how much} of that information is missing from the primary view. This two-axis design enables systematic evaluation of model performance across diverse occlusion conditions.
\begin{table}[t]
\centering
\small
\setlength{\abovecaptionskip}{2pt}
\begin{tabular}{lrrrr}
\toprule
Occlusion Type & Light & Medium & Heavy & Total \\
\midrule
Manipulated object & 257 & 412 & 231 & 900 \\
Receptacle & 186 & 338 & 226 & 750 \\
Dual & 57 & 250 & 43 & 350 \\
\midrule
\textbf{All} & \textbf{500} & \textbf{1000} & \textbf{500} & \textbf{2000} \\
\bottomrule
\end{tabular}
\caption{\textbf{Tasks distribution across occlusion types and severity levels in LIBERO-Occ.}}
\label{tab:libero_occ_stats}
\end{table}
\paragraph{Occlusion Types.}
We categorize occlusions according to the functional role of the occlusion target in the task. LIBERO-Occ includes three types.
\begin{itemize}[leftmargin=*,partopsep=0pt,topsep=0pt]
\setlength{\itemsep}{0pt}
\setlength{\parsep}{0pt}
\setlength{\parskip}{0pt}
    \item \textit{Manipulated object occlusion}: the object to be grasped, moved, or otherwise manipulated is occluded. This directly tests whether the model can recognize and localize the manipulation target under incomplete visual evidence.
    \item \textit{Receptacle occlusion}: the destination region, container, or contact area required for task completion is occluded. This evaluates whether the model can infer where an object should be placed or how it should interact when the goal region is not fully visible.
    \item \textit{Dual occlusion}: both the manipulated object and the receptacle are simultaneously occluded. This setting tests whether the model can jointly localize the object to act on and infer the partially visible goal region.
\end{itemize}

\paragraph{Occlusion Severity.}
Occlusion is not a binary attribute. When only a small region of the target is occluded, the model can rely on partial visual cues for localization. In more challenging scenarios where most of the target object is not visible, however, the model must infer its state from commonsense and spatial reasoning alone. To quantify occlusion severity, we compute the visibility loss of the occlusion target using instance segmentation. Let $A_{\mathrm{full}}$ denote the segmentation area of the target object in the original reference scene, and $A_{\mathrm{visible}}$ denote its area in the occluded observation. The occlusion severity score is defined as $S_{\mathrm{occ}} = (A_{\mathrm{full}} -A_{\mathrm{visible}})/A_{\mathrm{full}}$, measuring the fraction of the target object rendered invisible by occlusion.
Based on $S_{\mathrm{occ}}$, tasks are divided into three difficulty levels using per-suite quartiles: the bottom 25\% as light, the middle 50\% as medium, and the top 25\% as heavy. See Appendix~\ref{sec:appendix_occlusion_severity} for details of the split. This partition enables fine-grained analysis of how VLA performance changes as partial observability varies.

\paragraph{Benchmark Statistics.}
Table~\ref{tab:libero_occ_stats} reports the distribution of
tasks across occlusion types and severity levels. In total, LIBERO-Occ comprises 2{,}000 tasks spanning different occlusion target types, and severity levels, providing a comprehensive testbed for evaluating VLA robustness under realistic partial observability.

\begin{figure*}[t]
\setlength{\abovecaptionskip}{0pt}
    \centering
    \includegraphics[width=\linewidth]{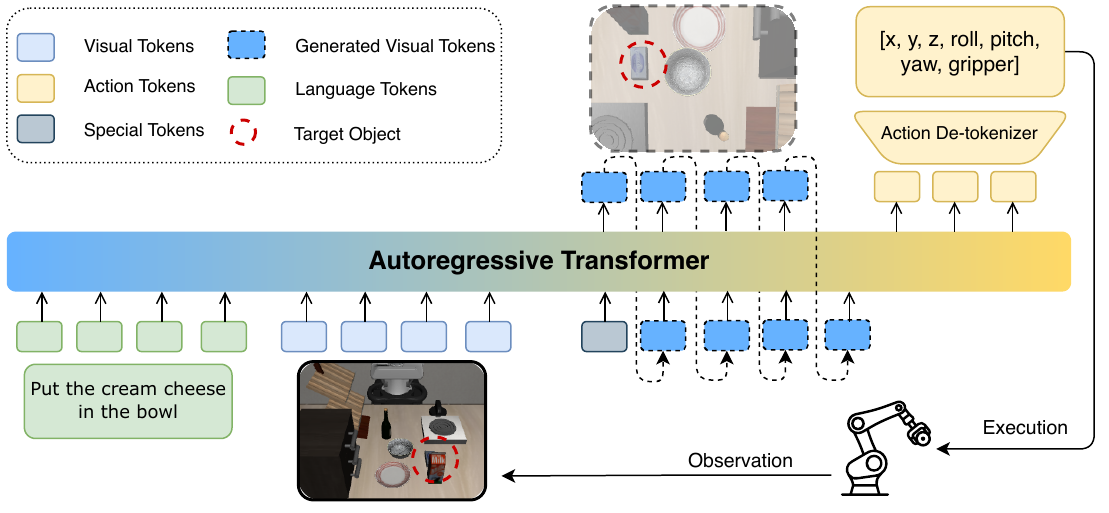}
    \caption{
\textbf{Viewpoint imagination framework.} Given a primary observation and a language instruction, the model uses an autoregressive transformer to first generate visual tokens of a complementary viewpoint and then predict action tokens. The imagined view may recover task-relevant information that is not visible from the primary camera, such as the shape, location, or surrounding context of an occluded object.
}
    \label{fig:framework}
\end{figure*}

\section{Methodology}
\subsection{Problem Formulation}
A standard VLA policy predicts robot actions directly from the current visual observation and language instruction. Given a primary view observation $o_t$, a task
instruction $l$, and an action chunk $a_t$, a vanilla VLA parameterized by $\theta$ models
\begin{equation*}
  p_\theta(a_t \mid o_t, l).
\end{equation*}
This formulation implicitly assumes that the observed image contains sufficient task-relevant spatial evidence for action prediction. However, under scene-induced occlusion, important entities may be partially hidden in $o_t$, making it difficult to infer well-grounded actions.
We therefore introduce a viewpoint imagination formulation. The model first predicts a complementary view $\hat{o}^{c}_t$ from the occluded primary observation, and then conditions action prediction on both the observed and imagined views:
\begin{equation*}
  p_\theta(\hat{o}^{c}_t, a_t \mid o_t, l) = \\
  p_\theta(\hat{o}^{c}_t \mid o_t, l)
  \cdot
  p_\theta(a_t \mid o_t, \hat{o}^{c}_t, l).
\end{equation*}
Here, $o_t$ is the available primary-view observation and $\hat{o}^{c}_t$ is a generated complementary view, such as a wrist or gripper view that may expose information occluded from the primary camera.

\subsection{Viewpoint Imagination Framework}
Figure \ref{fig:framework} illustrates our method. To jointly model imagined views and actions, the framework uses a unified autoregressive model that is capable of understanding and generating both textual and visual tokens. It consists of two coupled processes: viewpoint imagination and action prediction. First, given the occluded primary observation $o_t$ and the language instruction $l$, the model performs viewpoint imagination to generate the visual tokens of a complementary view, which serves as an intermediate representation of the potentially missing scene information.
The model then predicts action tokens conditioned on both the original observation and the generated complementary-view tokens.
In this way, the generated complementary view serves not merely as an auxiliary reconstruction target, but as an intermediate visual reasoning step that bridges perception under occlusion and policy generation. 

\subsection{Training and Inference}

\paragraph{Training} We train the model in two stages.
In the first stage, we train the model for viewpoint imagination. 
The model is given the occluded front view and the language instruction, and is optimized to generate the corresponding complementary view. This stage equips the model with the ability to infer hidden information from the occluded observation by learning cross-view spatial correspondence.

In the second stage, we jointly optimize viewpoint imagination and action prediction. The model predicts action tokens while continuing to receive supervision on complementary-view generation. The training objective is
  \begin{equation*}
  \mathcal{L} = 
  \mathcal{L}_{action} + 
  \lambda \mathcal{L}_{view}
  \end{equation*}
where $\mathcal{L}_{view} = - \sum_{i=1}^{N_v} \log p_\theta(v_i \mid v_{<i}, o_t, l)$ is the loss for complementary view generation
and $\mathcal{L}_{\mathrm{action}} = - \sum_{j=1}^{N_a} \log p_\theta(a_j \mid a_{<j}, \hat{o}^{c}_t, o_t, l)$ is for action prediction. We use group-wise loss weighting with $\lambda$ to balance the visual generation objective and the action prediction objective, as the two token groups can differ substantially in length. This stage teaches the model to use the inferred evidence for policy decisions.

\paragraph{Inference} At inference time, the robot receives only the current primary observation $o_t$ and language
instruction $l$. The model first generates an imagined complementary view $\hat{o}^{c}_t$. It then predicts the low-level action sequence $a_t$ conditioned on both the primary observation and the generated view. Since the complementary view is generated internally, the method can be deployed without additional cameras or active viewpoint control. If a real complementary view $o^{c}_t$ is available, it can directly replace the generated view. Therefore, our VIM framework supports both camera-free viewpoint imagination and camera-assisted deployment within a unified action-generation interface.

\section{Experiments}
\label{sec:experiments}

\subsection{Experimental Setup}
\paragraph{Benchmarks.}
We evaluate models on both the original LIBERO benchmark and our proposed LIBERO-Occ benchmark.
The original LIBERO benchmark provides a standard in-domain evaluation setting.
LIBERO-Occ, in contrast, introduces scene-induced occlusions while preserving the original task semantics and executable trajectory family. 
This enables us to isolate the effect of occlusion and examine whether VLA models can act reliably with missing information.
For the main evaluation, we report the average success rate over 500 rollouts for each task suite. For ablation studies, we use 100 rollouts per task suite to reduce evaluation cost.

%
\paragraph{Baselines.}
We compare VIM with representative state-of-the-art VLA models, including OpenVLA~\cite{kim2024openvla}, OpenVLA-OFT~\cite{Kim2025FineTuningVM}, $\pi$-0~\cite{black2024pi_0}, $\pi$-0.5~\cite{pmlr-v305-black25a}, and UniVLA~\cite{UniVLA}. These baselines cover both autoregressive and diffusion-based VLA policies and have shown strong performance on standard manipulation benchmarks.
\begin{table*}[t]
\centering
\setlength{\abovecaptionskip}{2pt}
\small
\setlength{\tabcolsep}{4.2pt}
\renewcommand{\arraystretch}{1.12}
\resizebox{\linewidth}{!}{
\begin{tabular}{lccccc|ccccc|c}
\toprule
\multirow{2}{*}{Method}
& \multicolumn{5}{c|}{Original LIBERO}
& \multicolumn{5}{c|}{LIBERO-Occ}
& \multirow{2}{*}{Avg. Drop $\downarrow$} \\
\cmidrule(lr){2-6} \cmidrule(lr){7-11}
& Avg. & Spatial & Goal & Object & LIBERO-10
& Avg. & Spatial & Goal & Object & LIBERO-10
&  \\
\midrule
UniVLA
& 88.25 & 86.00 & 90.00 & 95.00 & 82.00
& 57.10 & 61.00 & 67.40 & 78.80 & 21.20
& 31.15 \\
OpenVLA
& 92.65 & 94.00 & 95.20 & 96.40 & 85.00
& 40.65 & 39.20 & 37.40 & 65.60 & 20.40
& 52.00 \\
OpenVLA-OFT
& 95.75 & 97.00 & 93.00 & 97.00 & 93.00
& 47.95 & 48.00 & 44.80 & 75.60 & 23.40
& 47.80 \\
$\pi$-0
& 89.25 & 91.00 & 92.00 & 95.00 & 79.00
& 49.30 & 52.40 & 56.00 & 72.40 & 16.40
& 39.95 \\
$\pi$-0.5
& 90.00 & 90.00 & 90.00 & 97.00 & 83.00
& 40.55 & 39.00 & 40.20 & 64.00 & 19.00
& 49.45 \\
\midrule
Ours
& 90.75 & 94.00 & 91.00 & 98.00 & 80.00
& \textbf{65.05} & \textbf{76.40} & \textbf{74.60} & \textbf{84.20} & \textbf{25.00}
& \textbf{25.70} \\
\rowcolor{gray!12}
Ours w/ GT Comp.
& 93.00 & 93.00 & 92.00 & 98.00 & 89.00
& 74.00 & 83.00 & 81.00 & 92.00 & 40.00
& 19.00 \\
\bottomrule
\end{tabular}
}
\caption{
\textbf{Success rates on the original LIBERO and LIBERO-Occ benchmarks.}
We report results on each suite and the average over suites.
Avg. Drop denotes the decrease from the original LIBERO average to the LIBERO-Occ average.
Bold numbers indicate the best results under the complementary-view unavailable setting.
The gray row uses the ground-truth complementary view and is included only as a reference upper bound.
}
\label{tab:single_view_main}
\end{table*}
\paragraph{Implementation Details}
We implement our viewpoint imagination policy on top of the world model trained by UniVLA~\cite{UniVLA}, using an Emu3-MoE~\cite{wang2024emu3nexttokenpredictionneed} autoregressive backbone for unified image-token and action-token generation. Robot actions are tokenized by the FAST~\cite{pertsch2025fast} action tokenizer. For fair comparison, all methods are finetuned on the same demonstration data, preprocessed following~\cite{Kim2025FineTuningVM}. Training is conducted on 8 H100 GPUs with a batch size of 192. We use a cosine learning-rate schedule with a peak learning rate of $8\times10^{-5}$ for the first stage and $4\times10^{-5}$ for the second stage. We use the standard third-person view as the primary observation, and the wrist/gripper-camera view as the complementary view. We set $\lambda$ to $0.5$ in stage two. More implementation details are provided in appendix.
\paragraph{Observation Settings.}
We evaluate two observation settings. In the \textit{complementary-view available} setting, models receive both the primary view and the ground-truth complementary view, providing an upper-bound estimate when additional visual evidence is physically available. In the \textit{complementary-view unavailable} setting, models receive only the primary view, corresponding to the standard fixed-camera deployment scenario where task-relevant information may be occluded.

\begin{figure}[t]
\setlength{\abovecaptionskip}{0pt}
    \centering
    \includegraphics[width=\linewidth]{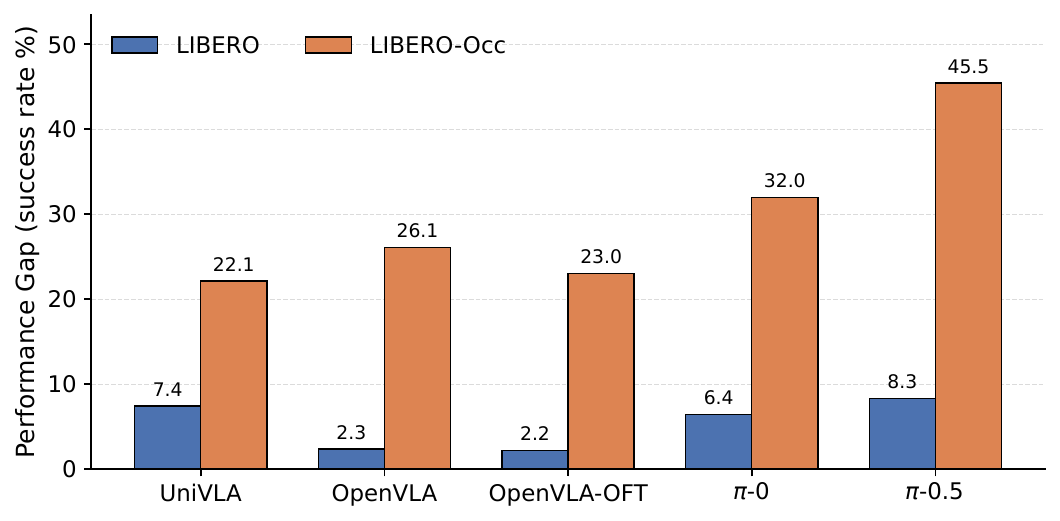}
    \caption{
    \textbf{Performance gap of VLA models when the complementary view is unavailable.}
    Larger drops indicate stronger dependence on additional visual evidence and weaker robustness under partial observability. Scene-induced occlusion substantially amplifies the benefit of complementary views, revealing that existing VLA models depend on additional visual evidence when task-relevant information is partially missing.
    }
    \label{fig:missing_view_drop}
\end{figure}

\subsection{Scene-Induced Occlusion Exposes Hidden Dependence on Complementary Views}
\label{sec:missing-view-results}
We first examine whether current VLA models rely on complementary visual information to maintain robust manipulation performance under occlusion. We report the performance gap between the complementary-view available setting and unavailable setting. As shown in Figure~\ref{fig:missing_view_drop}, the gain from providing the true complementary view is relatively modest on the original LIBERO benchmark, ranging from 2.2 to 8.3 percentage points across models. In contrast, the same gap increases substantially on LIBERO-Occ, ranging from 22.1 to 45.5 points. This sharp enlargement of the complementary-view gap indicates that scene-induced occlusion reveals a hidden dependence on viewpoints that expose task-relevant evidence. While standard LIBERO observations often contain sufficient information for action prediction, LIBERO-Occ turns manipulation into a partially observable problem, in which existing VLA policies still have limited ability to infer hidden scene states.

\begin{figure*}[t]
\setlength{\abovecaptionskip}{0pt}
    \centering
    \includegraphics[width=0.98\textwidth]{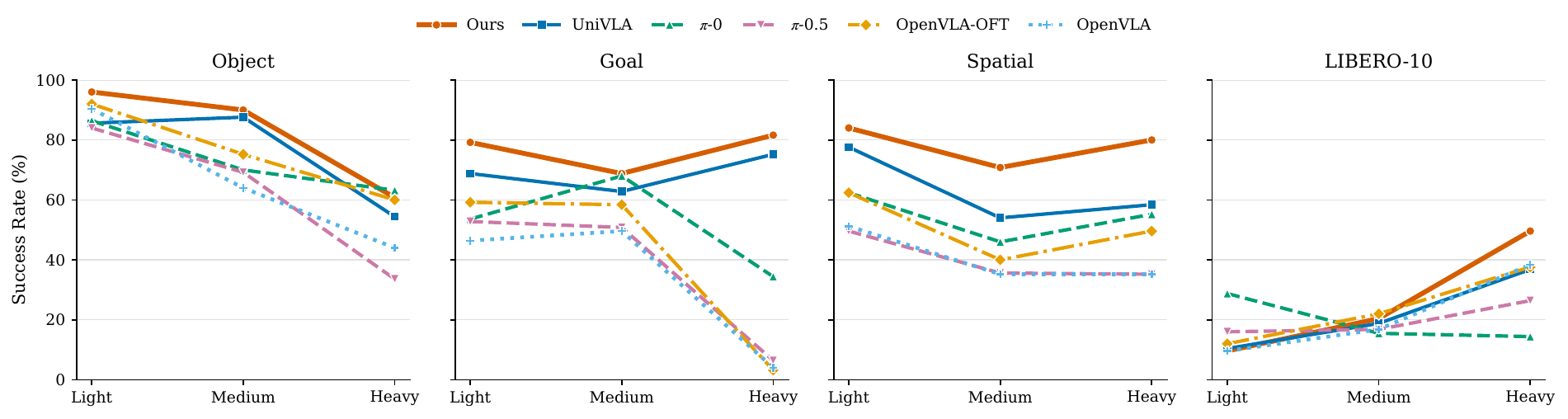}
    \caption{
    \textbf{Success rates under different occlusion severity levels on LIBERO-Occ.}
    We report results for Object, Goal, Spatial, and LIBERO-10 suites.
    As occlusion becomes more severe, existing VLA baselines generally suffer larger performance degradation, especially when task-relevant objects or goal regions are heavily hidden.
    Our method maintains stronger performance across severity levels, demonstrating improved robustness under increasing partial observability.
    }
    \label{fig:severity}
\end{figure*}
\subsection{Main Results on LIBERO and LIBERO-Occ}
\label{sec:main-results}

Table~\ref{tab:single_view_main} reports the results on both the original LIBERO and LIBERO-Occ benchmarks under complementary-view unavailable setting.  On the original LIBERO benchmark, all methods perform strongly, and OpenVLA-OFT achieves the best average success rate. However, this advantage does not transfer to occluded scenes: OpenVLA-OFT drops to 47.95\% on LIBERO-Occ, and other baselines exhibit similarly large declines. These results show that occlusion is a challenging and underexplored failure mode for current VLA systems. By contrast, our method achieves the highest LIBERO-Occ average success rate of 65.05\%, outperforming the strongest baseline by 7.95 percentage points. The improvement is consistent across different task suites, indicating that viewpoint imagination benefits different types of manipulation tasks rather than only a narrow subset. This result also supports the intuition that generative visual modeling can provide useful intermediate representations for recovering task-relevant evidence under occlusion.
Moreover, our method shows the smallest average drop from original LIBERO to LIBERO-Occ, suggesting that viewpoint imagination improves robustness to scene-induced occlusion without relying on additional cameras at deployment time. When the ground-truth complementary view is available, our model further improves the LIBERO-Occ average success rate from 65.05\% to 74.00\%, confirming that complementary visual evidence provides useful missing information. Qualitative examples in Appendix~\ref{sec:appendix_qualitative_examples} illustrate
  how the imagined complementary view recovers task-relevant spatial evidence in the occluded observation.

Overall, the main results show that LIBERO-Occ exposes a substantial robustness gap in current VLA models, while the proposed viewpoint imagination framework effectively mitigates this gap without requiring additional hardware at deployment time.

\subsection{Analysis across Occlusion Severity.}
We further analyze model performance under different occlusion severity levels, as shown in Figure~\ref{fig:severity}. Overall, stronger occlusion generally leads to larger performance degradation for existing VLA baselines. This confirms that LIBERO-Occ does not merely introduce superficial visual changes, but progressively hides task-relevant evidence needed for reliable action prediction. In the Object suite, most baselines drop sharply from light to heavy occlusion, indicating that directly hiding the object substantially weakens object recognition and localization. In the Goal suite, several baselines collapse under heavy occlusion, suggesting that inferring the destination region is particularly difficult when the target receptacle or placement area is no longer visible. In contrast, our method remains consistently competitive across severity levels and achieves the strongest overall performance in almost all settings. Notably, it preserves high success rates under heavy occlusions, where complementary spatial evidence is especially important. These results suggest that viewpoint imagination improves robustness under different levels of partial observability, by helping the policy recover missing information.

\subsection{Analysis by Occlusion Target Types.}
\begin{table}[t]
\setlength{\abovecaptionskip}{2pt}
\centering
\small
\setlength{\tabcolsep}{4.5pt}
\renewcommand{\arraystretch}{1.12}
\resizebox{\linewidth}{!}{
\begin{tabular}{lcccc}
\toprule
Method & Manipulated & Receptacle & Dual & Overall \\
\midrule
UniVLA & 47.78 & 81.87 & 28.00 & 57.10 \\
OpenVLA & 28.89 & 67.47 & 13.43 & 40.65 \\
OpenVLA-OFT & 35.22 & 77.87 & 16.57 & 47.95 \\
$\pi$-0 & 40.78 & 70.80 & 25.14 & 49.30 \\
$\pi$-0.5 & 29.78 & 67.87 & 9.71 & 40.55 \\
\midrule
Ours & \textbf{54.67} & \textbf{91.33} & \textbf{35.43} & \textbf{65.05} \\
\bottomrule
\end{tabular}
}
\caption{
\textbf{Success rates on LIBERO-Occ grouped by occlusion target type.}
Results are aggregated across all suites. The best result in each column is shown in bold.
}
\label{tab:occlusion_type_analysis}
\end{table}
We also analyze performance according to which task-relevant entity is occluded, as shown in Table~\ref{tab:occlusion_type_analysis}. The results reveal clear differences across occlusion types. Most models achieve higher success rates under receptacle occlusion. In contrast, manipulated object occlusion causes a larger degradation. This trend is also consistent with the asymmetric difficulty of manipulation behaviors: grasping a partially occluded object requires precise recognition and localization of the manipulated object, whereas placing an object into an occluded receptacle can often tolerate larger spatial uncertainty. Dual occlusion is the most challenging setting, as both the object to manipulate and the target region may be partially missing; all baselines drop substantially in this case. Our method achieves the best performance across all three occlusion types. These consistent gains indicate that viewpoint imagination is not merely exploiting an easier occlusion subset, but helps recover missing visual evidence and infer spatial relations across different forms of partial observability.

\subsection{Ablation Study}
\label{sec:ablation_two_stage}
\subsubsection{Ablation on Two-Stage Training}
We first study the role of the proposed two-stage training strategy. 
We compare three variants: 
(1) \textit{Ours w/o Stage-1 View Training}, which removes the viewpoint-imagination training stage and directly trains the model for action prediction with the Stage-2 objective; 
(2) \textit{Ours w/o Stage-2 View Loss}, which keeps Stage-1 but removes the auxiliary viewpoint-imagination loss during Stage-2; 
and (3) the full model.

\begin{table}[t]
\setlength{\abovecaptionskip}{2pt}
\centering
\small
\setlength{\tabcolsep}{3.5pt}
\renewcommand{\arraystretch}{1.10}
\begin{tabular}{lccccc}
\toprule
Variant & Avg. & Spatial & Goal & Object & 10 \\
\midrule
w/o S2 view loss 
& 0.00 & 0.00 & 0.00 & 0.00 & 0.00 \\
w/o S1 view train. 
& 36.25 & 42.00 & 39.00 & 61.00 & 3.00 \\
Full model 
& \textbf{65.00} & \textbf{78.00} & \textbf{73.00} & \textbf{84.00} & \textbf{25.00} \\
\bottomrule
\end{tabular}
\caption{
\textbf{Ablation study on the two-stage training strategy on LIBERO-Occ.}
S1: Stage-1 and S2: Stage-2.
}
\label{tab:ablation_two_stage}
\end{table}

Table~\ref{tab:ablation_two_stage} shows that both stages are essential for effective viewpoint imagination and downstream manipulation. 
Removing Stage-1 view training causes a substantial performance drop.
This indicates that directly learning action prediction without a dedicated viewpoint-imagination pretraining stage is insufficient for acquiring reliable cross-view spatial correspondence.

Removing the Stage-2 view loss causes complete failure across all suites. 
Inspection of the generated sequences shows that, the output tokens are interleaved with non-visual text and special tokens rather than forming a valid visual-token grid.
Because action prediction is autoregressively conditioned on this intermediate visual segment, this format collapse prevents reliable action generation. 
Thus, the Stage-2 view loss acts not merely as an auxiliary reconstruction objective, but as a structural regularizer that maintains a valid imagination-to-action generation interface.


\subsubsection{Ablation on Unified Viewpoint Imagination and Action Prediction}
\label{sec:ablation_unified_generation_policy}

We further evaluate whether unifying viewpoint imagination and action prediction is beneficial. 
To this end, we compare our full model with a separated pipeline, where our model first generates the complementary-view image and UniVLA then uses it for action prediction.
\begin{table}[t]
\setlength{\abovecaptionskip}{2pt}
\centering
\small
\setlength{\tabcolsep}{3.5pt}
\renewcommand{\arraystretch}{1.10}
\begin{tabular}{lccccc}
\toprule
Method & Avg. & Spatial & Goal & Object & 10 \\
\midrule
UniVLA
& 55.25 & 61.00 & 59.00 & 82.00 & 19.00 \\
Ours $\rightarrow$ UniVLA
& 62.00 & 72.00 & 69.00 & \textbf{84.00} & 23.00 \\
Ours
& \textbf{65.00} & \textbf{78.00} & \textbf{73.00} & \textbf{84.00} & \textbf{25.00} \\
\bottomrule
\end{tabular}
\caption{
\textbf{Ablation on unified viewpoint imagination and action prediction.}
UniVLA uses only the primary view.
"Ours $\rightarrow$ UniVLA" denotes the separated pipeline.
}
\label{tab:ablation_generated_view_univla}
\end{table}

As shown in Table~\ref{tab:ablation_generated_view_univla}, feeding our generated complementary view into UniVLA improves the average success rate by about 7 points. This indicates that the generated view contains task-relevant evidence that can also benefit an external VLA policy. Our unified model further improves the performance. The gap suggests that, by optimizing viewpoint imagination and action prediction within the same sequence-generation process, the unified autoregressive formulation helps the policy better exploit imagined visual tokens .

\section{Related Work}

\paragraph{Vision-Language-Action Models and Robust Manipulation Evaluation.}
Vision-Language-Action (VLA) models have shown strong potential for generalizable robotic manipulation by mapping visual observations and language instructions to robot actions~\cite{rt-2, palme, Kim2025FineTuningVM,octo,pmlr-v305-black25a}. They are commonly evaluated on benchmarks such as LIBERO~\cite{Liu2023LIBEROBK}, Bridge-V2~\cite{walke2023bridgedata}, and CALVIN~\cite{calvin}, where task-relevant objects and goal regions are usually visible from the input observation. Recent robustness studies further examine visual or environmental perturbations, including background changes, lighting variations, and image noise~\cite{libero-plus, libero-pro, robustVLA}. However, these settings often preserve task-relevant visual evidence. In contrast, scene-induced occlusion hides such evidence and makes manipulation partially observable. Our work isolates occlusion as a physically grounded evaluation factor and introduces LIBERO-Occ to systematically measure VLA robustness under occlusion.

\paragraph{Perception Completion for Manipulation under Occlusion.}
Occlusion is often addressed by acquiring additional visual evidence through multi-view perception or active camera control. Multi-camera robot policies use complementary viewpoints to provide useful spatial cues for manipulation~\cite{Bian2025VLALPAFLP,deng2025stereovla, reactive}, while active perception selects viewpoints that reduce uncertainty or reveal occluded objects~\cite{Asynchronous, Dai2025ActivePerceptiveLG}. Recent VLA-oriented methods further incorporate active viewpoint selection or task-aware view exploration into manipulation policies~\cite{Liu2026ActiveVLAIA, Liu2026SaPaVeTA}. Although effective, these approaches usually require additional cameras and calibration, increasing deployment cost and hardware complexity. In parallel, embodied world models and generative VLA models suggest that generative visual modeling can provide useful intermediate representations for control~\cite{ zhou2024robodreamer, cen2025worldvlaautoregressiveactionworld, UniVLA, zhao2025cot-vla}. VIM builds on this insight but uses generation for perception completion: it imagines a complementary view from the occluded primary observation and uses it as intermediate evidence for action prediction.

\section{Conclusion}

In this work, we study scene-induced occlusion as a practical partial-observability challenge for vision-language-action models. We introduce LIBERO-Occ, an occlusion-oriented extension of LIBERO. Experiments show that strong VLA models suffer substantial degradation under occlusion and rely more heavily on complementary views when task-relevant evidence is missing.

To address this challenge, we propose VIM, a viewpoint imagination framework that generates complementary visual evidence from an occluded primary observation and uses it for action prediction. Results on LIBERO-Occ show that VIM consistently improves robustness across task suites, severity levels, and occlusion target types without requiring additional cameras at inference time. These findings suggest that viewpoint imagination is a promising mechanism for perception completion in partially observable manipulation.

\section{Limitations}

This work has several limitations. First, LIBERO-Occ is constructed in simulation based on LIBERO. Although each occlusion is physically instantiated and verified through demonstration replay, the benchmark cannot fully capture the sensing noise, object variability, and dynamics of real-world manipulation environments. Thus, our results should be interpreted as a controlled study of scene-induced occlusion rather than a substitute for real-robot evaluation.

Second, VIM generates a complementary view from the primary observation, which makes it deployable without additional cameras but also constrains it by the model's learned visual prior. When the primary view contains very limited evidence, the imagined view may be inaccurate, especially under severe dual occlusion. Future work could improve reliability by modeling uncertainty or integrating temporal evidence across multiple frames.

Finally, our experiments mainly use the wrist/gripper view as the complementary viewpoint and require paired complementary-view observations during training. Other viewpoints, such as side, bird, or task-adaptive views, may provide different benefits, and such paired data may not always be available. Learning viewpoint imagination from weaker supervision or less structured video data remains an important direction for future work.
\bibliography{custom}

\appendix

\section{Details about LIBERO-Occ}
Following the original LIBERO protocol, each suite contains 10 task templates, and each task template is evaluated with 50 initial states that differ slightly in object configurations. In LIBERO-Occ, we instantiate occluders for these initial states. Because different occluder placements can induce different occlusion targets and severity levels, we treat each occluded initial state as a distinct occluded task instance. Thus, each LIBERO-Occ suite contains 500 occluded task instances, yielding 2,000 instances in total across the four suites.
\label{sec:appendix_occlusion_severity}
\subsection{Severity Distribution}
To characterize the occlusion severity distribution in each LIBERO-Occ suite, we report the first and third quartiles of the occlusion severity in Table \ref{tab:libero_occ_suite_quartiles}. The first quartile, $Q_1$, corresponds to the 25th percentile, meaning that 25\% of episodes have an
occlusion ratio no larger than this value. The third quartile, $Q_3$, corresponds to the
75th percentile, meaning that 75\% of episodes have an occlusion ratio no larger than this
value. These quartiles summarize the spread of occlusion severity across suites and define
the central range of occlusion ratios used in the benchmark.
\begin{table}[h]
  \centering
  \caption{Occlusion severity quartiles for each LIBERO-Occ suite. Severity is measured by the occlusion ratio in the manifest.}
  \label{tab:libero_occ_suite_quartiles}
  \begin{tabular}{lrrr}
  \toprule
  Suite & $N$ & $Q_1$ & $Q_3$ \\
  \midrule
Spatial & 500 & 0.5979 & 0.8874 \\
 Goal    & 500 & 0.3663 & 0.8239 \\
 Object  & 500 & 0.5771 & 0.8013 \\
  LIBERO-10      & 500 & 0.4278 & 0.8006 \\
  \bottomrule
  \end{tabular}
  \end{table}

\begin{figure}[h]
    \centering
    \setlength{\tabcolsep}{2pt}
    \captionsetup[subfigure]{font=small,skip=2pt}

    \begin{subfigure}[t]{0.48\linewidth}
        \centering
        \includegraphics[width=\linewidth]{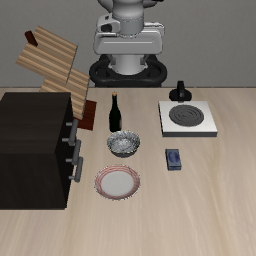}
        \caption{Original}
        \label{fig:appendix_severity_original}
    \end{subfigure}
    \hfill
    \begin{subfigure}[t]{0.48\linewidth}
        \centering
        \includegraphics[width=\linewidth]{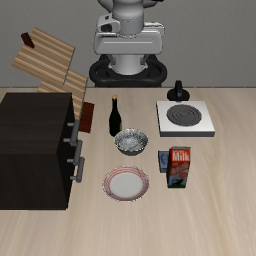}
        \caption{Light}
        \label{fig:appendix_severity_light}
    \end{subfigure}

    \vspace{0.3em}

    \begin{subfigure}[t]{0.48\linewidth}
        \centering
        \includegraphics[width=\linewidth]{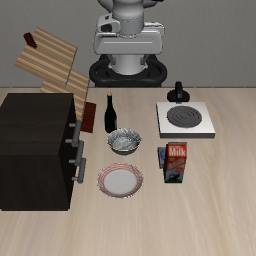}
        \caption{Medium}
        \label{fig:appendix_severity_medium}
    \end{subfigure}
    \hfill
    \begin{subfigure}[t]{0.48\linewidth}
        \centering
        \includegraphics[width=\linewidth]{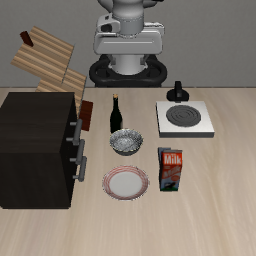}
        \caption{Heavy}
        \label{fig:appendix_severity_heavy}
    \end{subfigure}

    \caption{
    Examples of different occlusion severity levels in LIBERO-Occ.
    The original scene is shown for reference. As severity increases, more target-related visual evidence becomes unavailable from the primary view, making the task increasingly partially observable.
    }
    \label{fig:appendix_severity_examples}
\end{figure}
\subsection{Examples}
To provide an intuitive visualization of the severity split, Figure~\ref{fig:appendix_severity_examples} shows representative examples from the same task under different occlusion levels. Compared with the original scene, light occlusion only removes a small portion of the task-relevant visual evidence, while medium and heavy occlusions progressively hide larger parts of the target. These examples illustrate that the proposed severity metric reflects meaningful changes in partial observability rather than superficial appearance variations.

\section{Qualitative Examples}
  \label{sec:appendix_qualitative_examples}

\begin{figure*}[t]
      \centering
      \includegraphics[width=0.88\linewidth]{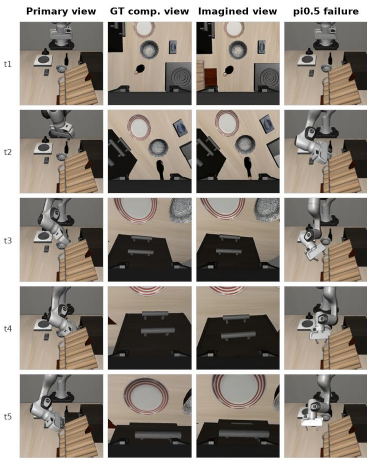}
      \caption{
      Qualitative example on LIBERO-Occ.
      Task instruction: \emph{``open the middle drawer of the cabinet''}.
      The drawer is partially occluded by scene objects. The imagined complementary view closely matches the ground-truth
      complementary view and provides additional spatial evidence for action prediction.
      In contrast, $\pi$-0.5 fails under the same primary-view setting.
      }
      \label{fig:qualitative_example_middle_drawer}
  \end{figure*}

\begin{figure*}[t]
      \centering
      \includegraphics[width=0.88\linewidth]{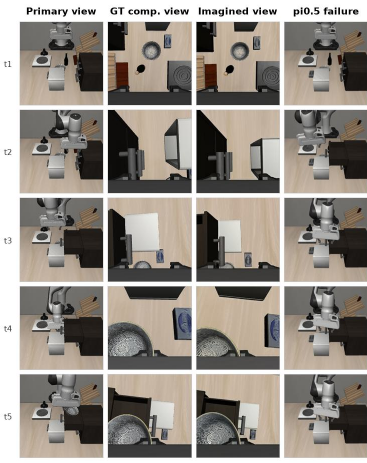}
      \caption{
      Qualitative example on LIBERO-Occ.
      Task instruction: \emph{``open the top drawer and put the bowl inside''}.
      The manipulated object becomes difficult to locate from the
      primary camera as the drawer, occluder, and robot arm block it.
      Our imagined complementary view recovers a viewpoint that exposes the drawer and bowl
      more clearly, supporting successful manipulation. $\pi$-0.5 fails on the same task when only the primary view is available.
      }
      \label{fig:qualitative_example_top_drawer}
  \end{figure*}

  We provide qualitative examples in Figures~\ref{fig:qualitative_example_middle_drawer}
  and~\ref{fig:qualitative_example_top_drawer}. Each figure shows five uniformly sampled
  timesteps from the rollout. The first three columns show the rollouts of our model, with columns correspond to the primary view, the
  ground-truth complementary view, the imagined complementary view generated.
  The fourth column is the primary-view rollout of $\pi$-0.5 on the same task. These examples illustrate that
  the imagined view provides task-relevant spatial evidence that is weakly visible or hidden
  from the primary camera, while the baseline policy fails under the same scene-induced
  occlusion.

\begin{table}[h]
  \centering
  \small
  \setlength{\tabcolsep}{5pt}
  \renewcommand{\arraystretch}{1.12}
  \begin{tabular}{ll}
  \toprule
  Setting & Value \\
  \midrule
  Base Model & UniVLA World Model \\
  Action tokenizer & FAST \\
  Hardware & 8 H100 GPUs \\
  Global batch size & 192 \\
  Precision & bfloat16 with TF32 enabled \\
  Optimizer & AdamW \\
  AdamW $\beta_1$, $\beta_2$ & 0.9, 0.95 \\
  Weight decay & 0.1 \\
  LR schedule & Cosine decay \\
  Stage-1 peak learning rate & $8\times10^{-5}$ \\
  Stage-2 peak learning rate & $4\times10^{-5}$ \\
  Stage-1 training steps & 4,000 \\
  Stage-2 training steps & 6,000 \\
  Stage-2 weight $
  \lambda$ & 0.5 \\
  Action chunk length & 10 \\
  Random seed & 42 \\
  Camera resolution & 200 \\
  \bottomrule
  \end{tabular}
  \caption{
  Implementation details for VIM training.
  }
  \label{tab:implementation_details}
  \end{table}
\section{Implementation Details}
  \label{sec:appendix_implementation}

  We summarize the main implementation details in Table~\ref{tab:implementation_details}. The model uses the third-person camera as the
  primary observation and the wrist/gripper camera as the complementary view.

\end{document}